\documentclass[11pt,a4paper]{article}
\usepackage[hyperref]{ranlp2025}
\usepackage{times}
\usepackage{latexsym}
\usepackage{amsmath,amssymb,amsfonts}
\usepackage{graphicx}
\usepackage{algorithm}
\usepackage{algorithmic}

\usepackage{microtype}
\usepackage{booktabs}
\usepackage{rotating}
\usepackage{multirow}
\aclfinalcopy 
\title{ILID: Native Script Language Identification for Indian Languages}
\author{Yash Ingle,
  Pruthwik Mishra \\
  Sardar Vallabhbhai National Institute of Technology,
  Surat, India \\
  yash.ingle003@gmail.com,pruthwikmishra@aid.svnit.ac.in
  }

\date{}

\begin{document}
\maketitle
\begin{abstract}
    The language identification task is a crucial fundamental step in NLP. Often it serves as a pre-processing step for widely used NLP applications such as multilingual machine translation, information retrieval, question and answering, and text summarization. The core challenge of language identification lies in distinguishing languages in noisy, short, and code-mixed environments. This becomes even harder in case of diverse Indian languages that exhibit lexical and phonetic similarities, but have distinct differences. Many Indian languages share the same script, making the task even more challenging. Taking all these challenges into account, we develop and release a dataset of 250K sentences consisting of 23 languages including English and all 22 official Indian languages labeled with their language identifiers, where data in most languages are newly created. We also develop and release baseline models using state-of-the-art approaches in machine learning and fine-tuning pre-trained transformer models. Our models outperforms the state-of-the-art pre-trained transformer models for the language identification task. The dataset and the codes are available at \url{https://yashingle-ai.github.io/ILID/} and in Huggingface open source libraries.
\end{abstract}
\section{Introduction}
India is a linguistically diverse country. Although there are more than 1000 languages in the Indian subcontinent, the digital divide among many Indian languages is enormous. Barring a few languages, most of the languages suffer from resource scarcity and a simple task such as Language IDentification (LID) remains a challenging task \cite{caswell-etal-2020-language} for them. Available LID tools such as \citet{lui-baldwin-2011-cross}, \citet{joulin2016fasttextzipcompressingtextclassification}, and \citet{nllbteam2022languageleftbehindscaling} perform poorly on Indian languages and they do not cover many of them. With the increase in mobile and internet users in India, the need for providing services in native Indian languages is more pressing than ever. Hence, it becomes essential to create robust LID tools to cater to Indian users. As the volume and variety of data continue to grow, we need to include data from different domains while creating sound benchmarks. Keeping this motivation in mind, we set out to create an Indian Language Identification (ILID) benchmark including English and 22 official languages spanning 25 scripts consisting of a total of 250K sentences labeled with their language markers. The main contribution of our paper is two fold. 
\begin{itemize}
    \item We create the ILID dataset consisting of newly created datasets in 13 languages and curated datasets in the remaining ten languages.
    \item We develop ILID baseline models using different machine learning and deep learning techniques.
    \item We perform a comparative analysis of our models with the state-of-the-art model.
\end{itemize}

\section{Related Work}

Language Identification (LID) is a fundamental task in Natural Language Processing (NLP), aiming to determine the language of a given text or speech segment. Early approaches to LID primarily relied on statistical methods, most notably character $n$-gram models \cite{Cavnar1994}. These methods leverage the unique frequency distributions of character sequences within different languages to classify unknown texts. Language Identification becomes more crucial in the domain of Cross Lingual Information Retrieval where the query is written in native scripts \cite{bosca2010language}. Similarly, spoken language identification\cite{dehak2011language,gonzalez2014automatic} is a very essential task in multilingual speech processing that can further impact in developing applications such as machine translation and speech recognition. In this paper, we are dealing with the language identification task in written texts. With machine learning approaches such as Support Vector Machines (SVMs) \cite{Dunning1994}, Naive Bayes classifiers \cite{elworthy-1998-language}, decision trees, random forests, and gradient boosting gaining widespread usage, LID systems were also modeled to leverage these techniques with the same $n$-gram features and other lexical or language specific hand crafted features. While effective for well-resourced languages with distinct orthographies, the performance of the statistical systems can degrade when dealing with short texts, noisy data, or languages sharing common scripts. Deep learning methods such as Convolutional Neural Networks (CNNs) \cite{Kim2014} and Recurrent Neural Networks (RNNs) \cite{Hochreiter1997} (particularly LSTMs and GRUs) are often trained on distributionally similar word or character n-gram or single character embeddings vectors to better capture intricate language complexity  and handle out-of-vocabulary words more robustly \cite{Conneau2017}. The current approaches include fine-tuning pre-trained transformer \cite{vaswani2017attention} models using subwords \cite{sennrich-etal-2016-neural,kudo-richardson-2018-sentencepiece,song-etal-2021-fast}. Many of the language identification tasks are modeled as token classification tasks that are helpful in code-mixed settings.

The task of language identification for Indian languages is even more challenging as they belong to different and diverse language families (Indo-Aryan, Dravidian, Tibeto-Burman, Austroasiatic), and written in over a dozen distinct scripts. Multiple Indian languages often share the same script. For example; Hindi, Marathi, Nepali, and Sanskrit are all primarily written in the Devanagari script. This script overlap requires models to distinguish languages based on lexical, morphological, or syntactic features rather than solely on orthographic cues.

Another prevalent issue in Indian language contexts is code-mixing and code-switching, where speakers or writers regularly use words or phrases from multiple languages within a single utterance or text \cite{Gamback2017}. This phenomenon is common in informal communication and social media \cite{barman2014code}, making it difficult for traditional LID systems to accurately identify the primary language or even segment code-mixed segments. Research in this area often focuses on fine-grained LID, aiming to identify language at the word or phrase level \cite{bhat2014iiit}, rather than just document or sentence level identification.

The language identification task for Indian languages also followed the similar line of research using rule-based and machine learning based approaches. More advanced approaches fine-tune pre-train transformers trained on Indian languages \cite{kumar-etal-2023-indisocialft,agarwal-etal-2023-limit,madhani-etal-2023-bhasa}. \citet{madhani-etal-2023-bhasa} also transliterate \cite{madhani-etal-2023-aksharantar} the native scripts into roman scripts for uniformity across languages and better efficiency in the LID systems. Despite these advancements, robust and fine-grained LID for all Indian languages, especially in code-mixed scenarios and for low-resource languages, remains an active area of research. In this paper, we limit the scope of identifying the language at a sentence level, not at the constituent token level.

\section{ILID Dataset Creation}
\label{sec:dataset}
Indian Language Identification (ILID) dataset is created using two approaches. We include English and the 22 official Indian languages \footnote{\url{https://en.wikipedia.org/wiki/Eighth_Schedule_to_the_Constitution_of_India}}\footnote{\url{https://en.wikipedia.org/wiki/Linguistic_Survey_of_India}} widely used in India. The first approach utilizes web scraping for the languages in which the digital presence is significant (details are given in the Appendix). For each of these languages, we collect 10,000 different sentences from various sources such as government websites, newspapers, books, and other public materials, ensuring varying degrees of linguistic complexity. The second approach samples sentences from Bhashaverse \cite{mujadia2024bhashaversetranslationecosystem}, an existing massive monolingual and parallel corpora for Indian languages. The details of the dataset are presented in Table~\ref{tab:language_data}. The data in each language is split into 80:10:10 ratio to create train, dev, and test sets. The dataset is available at \url{https://huggingface.co/datasets/yash-ingle/ILID_Indian_Language_Identification_Dataset}.
\begin{table*}[ht]
\centering
\begin{tabular}{lrrrr}
\toprule
\textbf{Language} & \textbf{\#Train} & \textbf{\#Dev} & \textbf{\#Test} & \textbf{\#Total} \\
\midrule
Assamese (asm) & 8000 & 1000 & 1000 & 10000 \\
Bengali (ben) & 8000 & 1000 & 1000 & 10000 \\
Bodo (brx) & 8000 & 1000 & 1000 & 10000 \\
Dogri (doi) & 8000 & 1000 & 1000 & 10000 \\
Konkani (gom) & 8000 & 1000 & 1000 & 10000 \\
Gujarati (guj) & 8000 & 1000 & 1000 & 10000 \\
Hindi (hin) & 8000 & 1000 & 1000 & 10000 \\
Kannada (kan) & 8000 & 1000 & 1000 & 10000 \\
Kashmiri (kas) & 8000 & 1000 & 1000 & 10000 \\
Maithili (mai) & 8000 & 1000 & 1000 & 10000 \\
Malayalam (mal) & 8000 & 1000 & 1000 & 10000 \\
Marathi (mar) & 8000 & 1000 & 1000 & 10000 \\
Manipuri Bengali Script (mni\_Beng) & 8000 & 1000 & 1000 & 10000 \\

Manipuri Meitei Script (mni\_Mtei) & 8000 & 1000 & 1000 & 10000 \\
Nepali (npi) & 8000 & 1000 & 1000 & 10000 \\
Odia (ory) & 8000 & 1000 & 1000 & 10000 \\
Punjabi (pan) & 8000 & 1000 & 1000 & 10000 \\
sanskrit(san)&8000  &1000  &1000  & 10000\\
Santali (sat) & 8000 & 1000 & 1000 & 10000 \\
Sindhi Devnagari(snd\_Deva) & 8000 & 1000 & 1000 & 10000 \\
Sindhi Perso-Arabic(snd\_Arab) & 8000 & 1000 & 1000 & 10000 \\
Tamil (tam) & 8000 & 1000 & 1000 & 10000 \\
Telugu (tel) & 8000 & 1000 & 1000 & 10000 \\
Urdu (urd) & 8000 & 1000 & 1000 & 10000 \\
English (eng) & 8000 & 1000 & 1000 & 10000 \\
\midrule
\textbf{Total} & \textbf{200000} & \textbf{25000} & \textbf{25000} & \textbf{250000} \\
\bottomrule
\end{tabular}
\caption{Data splits for ILID Benchmark Dataset}
\label{tab:language_data}
\end{table*}

\begin{figure*}[ht]
\hspace*{0pt}
\includegraphics[width=1.1\textwidth]{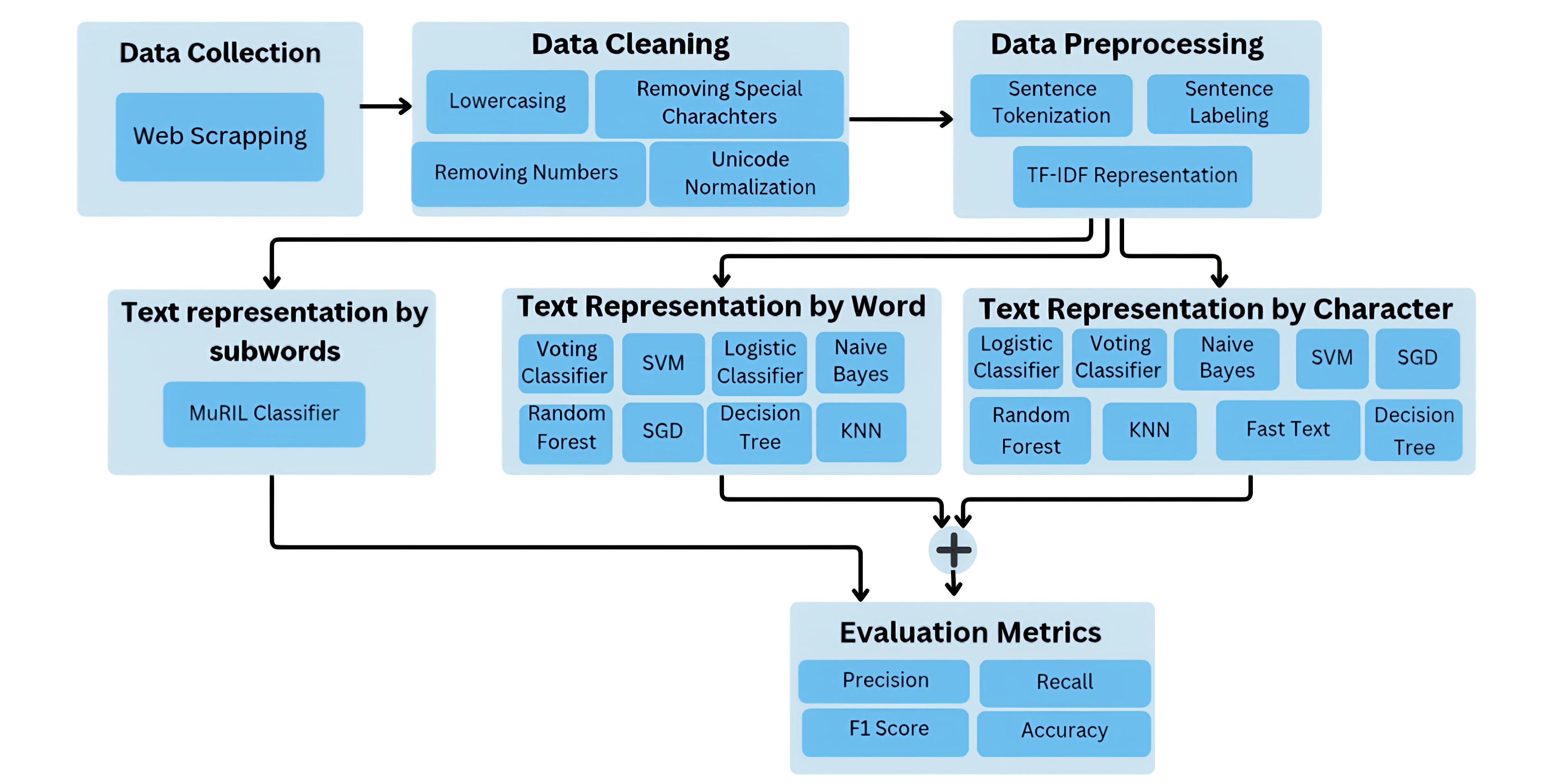}
\caption{Workflow of the Proposed Language Identification System.}
\label{fig:workflow}
\end{figure*}

To ensure quality and consistency, the dataset undergoes several noise removal steps. The first step involves the elimination of duplicate, very short, and ungrammatical sentences. The next step employs an existing FastText \cite{joulin2016bagtricksefficienttext,joulin2016fasttextzipcompressingtextclassification} based language identification model to remove sentences where the probability of detecting the language of the text is very low (we fix a threshold of 0.7 for this filtering process). This happens in code-mixed texts as many Indians are multilingual
and often use more than one language while writing.

The approaches are described in detail.
\subsection{Web Scraping}
To build a strong language identification system for Indian languages, we carefully built a personalized set of 10,000 text samples for each language, covering 13 Indian languages.

Text data is collected from a variety of publicly available and diverse sources, including Wikipedia dumps in respective languages for formal and structured text, news websites and blogs for professional and personal updates. All efforts are made to ensure the samples are multilingual in length, script (i.e., Devanagari, Bengali, Tamil, Odia etc.), domain, and style, reflecting the multilingual nature of India.

For a target website $W$, we define the scraping process as:

\begin{equation}
    S(W) = \bigcup_{p \in P_W} \phi(p)
\end{equation}

where $P_W$ is the set of target pages and $\phi$ is the extraction function that parses HTML content while preserving structural information. We implement adaptive throttling using:

\begin{equation}
    \Delta_t = \frac{1}{|W|} \sum_{i=1}^{|W|} \frac{\text{size}(w_i)}{\text{bandwidth}}
\end{equation}

\subsection{Data Cleaning Pipeline}
In the data cleaning process, we mainly work on removing extra spaces, extra special symbols, unicode normalization, breaking the paragraph into meaningful sentences, and after that we apply various tokenization methods. For sentence tokenization, we utilize the corresponding end sentence markers defined for respective languages. We use a regular expression based tokenizer \footnote{\url{https://github.com/Pruthwik/Tokenizer_for_Indian_Languages}} for sentence and word tokenization  that is specifically designed to handle texts in Indian languages.
Our cleaning process transforms raw text $T$ to cleaned text $T'$ through:
\vspace{-1.24em}
\begin{equation}
    T' = \psi_n(\psi_s(T))
\end{equation}

where $\psi_s$ handles special characters and $\psi_n$ performs normalization. We use a sentence as a unit for the LID task. Each sentence is labeled with a language tag. ISO 639-2 codes \footnote{\url{https://en.wikipedia.org/wiki/List_of_ISO_639-2_codes}} or 3 lettered language codes denote each language detailed in Table~\ref{tab:lang_code_iso}. 
\subsection{Sampling From Existing Corpora}
In the second approach, sentences are randomly sampled from Bhashaverse \cite{mujadia2024bhashaversetranslationecosystem} that contain a huge collection of monolingual corpora of different Indian languages. In this corpora, Manipuri, and Sindhi have sentences in two different scripts each. Manipuri is available in Bengali and Meitei scripts, whereas Sindhi is written in Devanagari and Perso-Arabic scripts. We have sampled sentences using all available scripts from these two languages. This increases the total number of languages with different scripts to 25 also presented in Table~\ref{tab:corpus_stats}. A similar exercise of appending the labels to the sentences as language identifiers and data splitting is performed on these samples.
The language specifics using both approaches are shown in Table~\ref{tab:langs}.

\begin{table}[ht]
\centering
\begin{tabular}{|l|l|}
\hline
\textbf{Scraping} & \textbf{Sampling} \\ \hline
Assamese          & Bodo              \\ \hline
Bengali           & Dogri             \\ \hline
Gujarati          & Kashmiri          \\ \hline
Hindi             & Konkani           \\ \hline
Kannada           & Maithili          \\ \hline
Malayalam         & Manipuri          \\ \hline
Marathi           & Nepali            \\ \hline
Oriya             & Sanskrit          \\ \hline
Punjabi           & Santhali          \\ \hline
Tamil             & Sindhi            \\ \hline
Telugu            &                   \\ \hline
Urdu              &                   \\ \hline
English           &                   \\ \hline
\end{tabular}
\caption{Languages Used in Scraping and Sampling}
\label{tab:langs}
\end{table}

\begin{table}[ht]
\centering
\begin{tabular}{|c|c|}
\hline
\textbf{Language} & \textbf{ISO 639-2 Code} \\ \hline
Assamese          & asm                     \\ \hline
Bangla            & ben                     \\ \hline
Bodo              & brx                     \\ \hline
Dogri             & doi                     \\ \hline
Gujarati          & guj                     \\ \hline
Hindi             & hin                     \\ \hline
Kannada           & kan                     \\ \hline
Kashmiri          & kas                     \\ \hline
Konkani           & gom                     \\ \hline
Maithili          & mai                     \\ \hline
Malayalam         & mal                     \\ \hline
Manipuri Bengali Script& mni\_Beng          \\ \hline
Manipuri Meitei Scrip& mni\_Mtei            \\ \hline
Marathi           & mar                     \\ \hline
Nepali            & npi                     \\ \hline
Oriya             & ory                     \\ \hline
Punjabi           & pan                     \\ \hline
Sanskrit          & san                     \\ \hline
Santali           & sat                     \\ \hline
Sindhi Perso-Arabic&snd\_Arab               \\ \hline
Sindhi Devanagari & snd\_Deva                     \\ \hline
Tamil             & tam                     \\ \hline
Telugu            & tel                     \\ \hline
Urdu              & urd                     \\ \hline
English           & eng                     \\  \hline
\end{tabular}
\caption{Language Full Name to ISO 639-2 Language Codes Mapping}
\label{tab:lang_code_iso}
\end{table}

\begin{table*}[ht]
\centering
\renewcommand{\arraystretch}{0.9}
\resizebox{1.\textwidth}{!}{
\begin{tabular}{llllllll}
\toprule
\textbf{lang} & \textbf{\#sents} & \textbf{\#words} & \textbf{\#chars} & \textbf{avg\_word\_len} & \textbf{avg\_sent\_len} & \textbf{\#unique\_words} & \textbf{TTR\_words} \\\midrule
mai       & 10000   & 65884   & 288223  & 3.526          & 28.822         & 6743     & 0.102      \\
eng       & 10000   & 225036  & 1052283 & 3.721          & 105.228        & 22709    & 0.101      \\
kan       & 10000   & 122400  & 861095  & 6.117          & 86.109         & 34644    & 0.283      \\
san       & 10000   & 130353  & 858829  & 5.665          & 85.883         & 50506    & 0.387      \\
gom       & 10000   & 58282   & 296807  & 4.264          & 29.681         & 8104     & 0.139      \\
hin       & 10000   & 106402  & 476919  & 3.576          & 47.692         & 8451     & 0.079      \\
mar       & 10000   & 89232   & 526021  & 5.007          & 52.602         & 19470    & 0.218      \\
asm       & 10000   & 64251   & 311313  & 4.001          & 31.131         & 7196     & 0.112      \\
urd       & 10000   & 97233   & 387031  & 3.083          & 38.703         & 19987    & 0.206      \\
tel       & 10000   & 81281   & 524216  & 5.572          & 52.422         & 19554    & 0.241      \\
tam       & 10000   & 144307  & 1018268 & 6.126          & 101.827        & 45912    & 0.318      \\
snd\_Arab & 10000   & 113249  & 479446  & 3.322          & 47.945         & 7409     & 0.065      \\
ben       & 10000   & 170048  & 877189  & 4.217          & 87.719         & 30007    & 0.176      \\
snd\_Deva & 10000   & 70743   & 312570  & 3.56           & 31.257         & 6249     & 0.088      \\
ory       & 10000   & 91919   & 526927  & 4.841          & 52.693         & 5400     & 0.059      \\
npi       & 10000   & 59095   & 296473  & 4.186          & 29.647         & 7966     & 0.135      \\
mni\_Mtei & 10000   & 73043   & 318189  & 3.493          & 31.819         & 7245     & 0.099      \\
mal       & 10000   & 117447  & 985898  & 7.48           & 98.59          & 37010    & 0.315      \\
doi       & 10000   & 75469   & 313613  & 3.288          & 31.361         & 5904     & 0.078      \\
pan       & 10000   & 168450  & 721285  & 3.341          & 72.129         & 18032    & 0.107      \\
brx       & 10000   & 65717   & 343731  & 4.383          & 34.373         & 7828     & 0.119      \\
sat       & 10000   & 78554   & 360814  & 3.72           & 36.081         & 4067     & 0.052      \\
mni\_Beng & 10000   & 186109  & 1208020 & 5.545          & 120.802        & 27350    & 0.147      \\
guj       & 10000   & 175966  & 911173  & 4.235          & 91.117         & 28195    & 0.16       \\
kas       & 10000   & 69270   & 297326  & 3.437          & 29.733         & 7752     & 0.112     \\\bottomrule
\end{tabular}
}
\caption{Language Wise Corpus Statistics Where the column TTR\_words refers to the Type Token Ratio for Words}
\label{tab:corpus_stats}
\end{table*}

\begin{table*}[ht]
\centering
\renewcommand{\arraystretch}{0.9} 
\begin{tabular}{|c|c|c|c|c|c|c|c|cc|c|}
\hline
\multirow{2}{*}{\textbf{Model}} & \multirow{2}{*}{\textbf{KNN}} & \multirow{2}{*}{\textbf{DT}} & \multirow{2}{*}{\textbf{RF}} & \multirow{2}{*}{\textbf{SVM}} & \multirow{2}{*}{\textbf{NB}} & \multirow{2}{*}{\textbf{LogReg}} & \multirow{2}{*}{\textbf{SGD}} &
\multicolumn{2}{c|}{\textbf{ILID}} & \multirow{2}{*}{\textbf{Bh-Ab}}\\
&&&&&&&&Dev&Test&\\\hline

Voting-1 & \checkmark & \checkmark & \checkmark &  & \checkmark & \checkmark &  & \textbf{1.00} & 0.96 &\textbf{1.00} \\\hline
Voting-2 & \checkmark & \checkmark & \checkmark &  &  & \checkmark &\checkmark  &\textbf{1.00} & 0.96&\textbf{1.00} \\\hline
Voting-3 &  & \checkmark &\checkmark  &  &\checkmark  & \checkmark & \checkmark & \textbf{1.00}& 0.96 &\textbf{1.00} \\\hline 
Voting-4 & \checkmark & \checkmark & \checkmark & \checkmark & \checkmark &  &  & 0.99 & 0.96 &\textbf{1.00} \\\cline{1-11} 
Voting-5 &  &  & \checkmark & \checkmark & \checkmark & \checkmark & \checkmark & 0.99 & \textbf{0.99} &\textbf{1.00}\\\hline
MuRIL    &  &  &  &  &  &  &  & 0.96 & 0.96&0.9 \\\hline
FastText &  &  &  &  &  &  &  & 0.96 & 0.96&0.9 \\
\hline
\end{tabular}
\caption{Comparison of Voting Classifiers, MuRIL, and FastText on ILID dataset and Bhasha-Abhijnaanam (Bh-Ab) dataset. Highest scores are marked in bold.}
\label{tab:best_scores_own}
\end{table*}

\subsection{Corpora Statistics}
After collecting and cleaning the dataset, we compute several corpora statistics. These statistics are excellent indicators of the levels and complexities of each language. The properties of the corpora considered include the total number of words, the total number of characters, the total number of unique words, average sentence length, the average word length, and the type / token ratio \cite{textinspector} which are markers of the representativeness of the corpora. These statistics are shown in Table~\ref{tab:corpus_stats}.
\section{ILID Model}
\label{sec:approach}

The ILID model is a classifier that can categorize a piece of text into one of the 25 classes that includes English and 22 Indian languages. The architecture of the proposed system is shown in Figure~\ref{fig:workflow}. We implement three types of approach for designing the classifiers.
\subsection{Machine Learning Models}
In this approach, each sentence is represented by a TF-IDF \cite{sparck1972statistical} vector in a bag-of-words setting. We utilize both word-level and character-level TF-IDF representations to enhance textual feature extraction. 
\subsubsection{Word-Level TF-IDF Based Classification}
We begin by utilizing word-level TF-IDF, computed using uni-grams and bi-grams. Higher word n-grams($n>2$) are not used for modeling TF-IDF as they suffer from sparsity. This representation captures the importance of whole words and adjacent word pairs across the corpus, allowing the model to understand the semantic context and overall meaning of sentences. However, word-based models are sensitive to spelling variations and may not generalize well to morphologically rich or noisy text, which is common in Indian languages.

\subsubsection{Character-Level TF-IDF Based Classification}
To overcome the limitations of word-level models, we introduce character-level TF-IDF representations using character n-grams ranging from 2 to 6. This approach captures sub-word structures, prefixes, suffixes, and root forms that are particularly effective in handling typographical errors, morphological variations, and out-of-vocabulary words. Character-level features are especially helpful in language identification tasks involving noisy or informal text, enabling the model to learn finer linguistic patterns.

\subsubsection{Combined Feature-Based Classification}

To leverage the strengths of both word and character level features, we develop a combined TF-IDF representation by concatenating the two feature sets into a single vector. This hybrid approach provides a more robust and language-agnostic representation. We train eight traditional machine learning classifiers—Support Vector Machine (SVM), Logistic Regression (LR), Random Forest (RF), Decision Tree (DT), Naive Bayes (NB), Stochastic Gradient Descent (SGD), and K-Nearest Neighbors (KNN)—on these features. Furthermore, we explore various ensemble models involving combinations of 3, 4, and 5 diverse base classifiers from the pool of classifiers. Due to space constraints, only the top five performing ensembles are selected based on F1 scores on the development and test sets, are presented in Table~\ref{tab:best_scores_own}.

\subsection{FastText Classifier}
FastText \cite{joulin2016bagtricksefficienttext,joulin2016fasttextzipcompressingtextclassification} utilizes word embeddings \cite{bojanowski2017enriching} composed of character n-grams that better represent rare words and orthographic similarities. FastText classifier is a linear classifier and is very fast in computation. It can generate embeddings for out-of-vocabulary (OOV) words, which are common in user-generated or noisy text data.
\subsection{Pretrained BERT Model}
Pretrained subword based contextual language models such as BERT \cite{devlin2019bert}, XLM \cite{conneau-etal-2020-unsupervised}, RoBERTa \cite{zhuang-etal-2021-robustly} have proven to be very effective in text classification tasks or generally natural language understanding tasks. One variant of BERT, MuRIL \cite{khanuja2021muril} is pretrained on large amounts of corpora in Indian languages. The pre-training data is also augmented with translated and transliterated texts that makes the model capture cross-lingual and code-mixed embeddings more efficiently. Hence, we fine-tune the MuRIL pretrained model on the ILID train set for our task.
\section{Experimental Details}
The machine learning models have been implemented using the Scikit-learn \cite{pedregosa2011scikit} framework. Similarly, the FastText library \footnote{\url{https://fasttext.cc/docs/en/supervised-tutorial.html}} is utilized to implement the language identifiers of the texts modeled as text classifiers \cite{joulin2016bagtricksefficienttext,joulin2016fasttextzipcompressingtextclassification}. The MuRIL \cite{khanuja2021muril} models are fine-tuned using the Huggingface transformer \cite{wolf2019huggingface} framework on an NVIDIA H100 GPU with 94GB RAM. The batch size, the number of epochs, the learning rate, maximum sequence length, and weight decay are set to 32, 10, 0.00002, 256, and 0.01 respectively. The fine-tuned MuRIL model is available at \url{https://huggingface.co/pruthwik/ilid-muril-model}.

\section{Evaluation Metrics}

All models have been evaluated using the macro F1 scores. The macro F1 score averages the F-1 scores across all languages. In order to evaluate the score the F1-score of each language, the precision and recall scores are also computed at the macro level. Each ensemble employs a voting mechanism to determine the predicted labels. Two types of voting mechanism are used: soft and hard. Hard voting is based on majority voting among the classifiers present in the ensemble. This type of voting is performed when the SVM classifier is part of the ensemble because SVM is inherently a non-probabilistic classifier. Soft voting aggregates the probability estimates of individual classifiers in the ensemble and the class with argmax of the sums of the predicted probabilities is chosen as the label. Ensembles where SVM is not included are evaluated using soft voting.

\section{Results and Discussion}

The ensemble machine learning models perform better than the individual models. The performance of the ensembles is also superior to the FastText and fine-tuned MuRIL models. The best five performing ensembles along with the FastText and fine-tuned MuRIL models are presented in Table~\ref{tab:best_scores_own} that represent the macro F1 score for each model. 
Although the scores are encouraging, the performance drops in languages that share a script such as Devanagari for Hindi, Maithili, Marathi, Konkani, and Sanskrit, Arabic for Kashmiri, Urdu, and Sindhi. We compare our models with the state-of-the-art IndicLID model \cite{madhani-etal-2023-bhasa} on the Bhasha-Abhijnaanam \cite{madhani-etal-2023-bhasa} benchmark of 88K sentences where our model outperforms the IndicLID model \cite{madhani-etal-2023-bhasa}. The F1-scores of the IndicLID model on the Bhasha-Abhijnaanam dataset is 0.98 as reported in the paper. We could not run their fine-tuned IndicBERT model while the IndicLID FTN model has a macro F1-score of 0.88. Our deep learning baselines perform better on the languages where data are scraped, while the performance dips for languages that are created from other external resources.
\subsection{Results for Individual Languages}
The language-wise performance of the models is shown in Tables~\ref{tab:combined-f1-table}. We can observe from the table that the performance drops for extremely low resource languages such as Bodo, Dogri, Maithili, and Sindhi. Deep learning models suffer in identifying those languages for which they have scarcity during pre-training. MuRIL \cite{khanuja2021muril} outperforms every other model in languages on which the model was pre-trained.
\section{Conclusion}
In this paper, we present the ILID benchmark, a manually curated dataset for 23 languages, which includes English and 22 Indian languages containing 250K sentences. It is specifically designed for Indian language identification. Along with the dataset, we develop several machine learning an deep learning models that perform well, even with limited training data. This makes them especially suitable for low-resource languages. To check the consistency and reliability of our ML models, we compare them with deep learning models like MuRIL and FastText. The ILID dataset and models together provide important resources for advancing multilingual NLP research in Indian languages. We hope this work encourages more exploration and development in this area, which has been overlooked.


\begin{table*}[ht]
\centering
\renewcommand{\arraystretch}{2.0}
\resizebox{\textwidth}{!}{%
\begin{tabular}{|c|cc|cc|cc|cc|cc|cc|cc|cc|cc|}
\hline
\multirow{2}{*}{\textbf{Lang}} & \multicolumn{2}{c|}{\textbf{LR}} & \multicolumn{2}{c|}{\textbf{DT}} & \multicolumn{2}{c|}{\textbf{RF}} & \multicolumn{2}{c|}{\textbf{SVM}} & \multicolumn{2}{c|}{\textbf{SGD}} & \multicolumn{2}{c|}{\textbf{KNN}} & \multicolumn{2}{c|}{\textbf{NB}}   & \multicolumn{2}{c|}{\textbf{Muril}}  & \multicolumn{2}{c|}{\textbf{Fasttext}} \\
 & Dev & Test & Dev & Test & Dev & Test & Dev & Test & Dev & Test & Dev & Test & Dev & Test & Dev & Test & Dev & Test\\
\hline
asm       & 0.98 & \textbf{0.99} & 0.96 & 0.96 & 0.98 & 0.98 & 0.98 & \textbf{0.99} & 0.97 & 0.98 & 0.90 & 0.96 & 0.97 & 0.98  & \textbf{0.99} & \textbf{0.99} & 0.96 & 0.96\\
ben       & \textbf{1.00} &\textbf{1.00} & 0.97 & 0.98 & 0.99 & 0.99 & \textbf{1.00} & \textbf{1.00} & 0.99 & 0.99 & 0.91 & 0.97 & 0.99 & 0.99 & \textbf{1.00} & \textbf{1.00} & 0.98 & 0.98\\
brx       & \textbf{0.96} & \textbf{0.95} & 0.91 & 0.89 & \textbf{0.96} & 0.94 & \textbf{0.96} & \textbf{0.95} & 0.94 & \textbf{0.95} & 0.85 & 0.93 & \textbf{0.96} & \textbf{0.95} & 0.91 & 0.91 & 0.92 & 0.91\\
doi       & 0.92 & \textbf{0.92} & 0.83 & 0.82 & 0.90 & 0.91 & 0.91 & \textbf{0.92} & 0.88 & 0.88 & 0.68 & 0.81 & 0.90 & 0.91 & \textbf{0.93} & \textbf{0.92} & 0.91 & 0.91\\
eng       & \textbf{1.00} & \textbf{1.00} & 0.99 & 0.99 & \textbf{1.00} & \textbf{1.00} & \textbf{1.00} & \textbf{1.00} & \textbf{1.00} & \textbf{1.00} & 0.99 & \textbf{1.00} & \textbf{1.00} & 0.99 & \textbf{1.00} & \textbf{1.00} & 0.99 & 0.99\\
gom       & \textbf{0.97} & \textbf{0.97} & 0.84 & 0.83 & 0.94 & 0.95 & 0.96 & \textbf{0.97} & 0.90 & 0.91 & 0.78 & 0.90 & 0.94 & 0.96 & \textbf{0.97} & \textbf{0.97} & 0.91 & 0.92\\
guj       & \textbf{1.00} & \textbf{1.00} & \textbf{1.00} & \textbf{1.00} & \textbf{1.00} & \textbf{1.00} & \textbf{1.00} & \textbf{1.00} & \textbf{1.00} & \textbf{1.00} & \textbf{1.00} & \textbf{1.00} & 0.95 & \textbf{1.00} & \textbf{1.00} & \textbf{1.00} & 0.99 & 0.99\\
hin       & 0.99 & 0.98 & 0.93 & 0.92 & 0.98 & 0.97 & 0.99 & \textbf{0.99} & 0.97 & 0.97 & 0.80 & 0.92 & \textbf{1.00} & 0.95 & 0.99 & \textbf{0.99} & 0.98 & 0.98\\
kan       & \textbf{1.00} & \textbf{1.00} & \textbf{1.00} & \textbf{1.00} & \textbf{1.00} & \textbf{1.00} & \textbf{1.00} & \textbf{1.00} & \textbf{1.00} & \textbf{1.00} & \textbf{1.00} & \textbf{1.00} & 0.98 & \textbf{1.00} & \textbf{1.00} & \textbf{1.00} & 0.98 & 0.97\\
kas       &\textbf{0.99} & 0.98 & 0.94 & 0.94 & 0.97 & 0.98 & \textbf{0.99} & 0.99 & 0.94 & 0.95 & 0.96 & 0.96 & 0.85 & 0.98  & \textbf{0.99} & \textbf{1.00} & 0.97 & 0.96\\
mai       & 0.90 & 0.86 & 0.83 & 0.78 & 0.84 & 0.84 & \textbf{1.00} & 0.85 & 0.83 & 0.82 & 0.65 & 0.73 & 0.84 & 0.85  & 0.89 & \textbf{0.88} & 0.87 & 0.86\\
mal       & \textbf{1.00} & \textbf{1.00} & 0.99 & \textbf{1.00} & \textbf{1.00} & \textbf{1.00} & 0.86 & \textbf{1.00} & \textbf{1.00} & \textbf{1.00} & \textbf{1.00} & \textbf{1.00} & 0.95 & \textbf{1.00}  & \textbf{1.00} & \textbf{1.00} & 0.94 & 0.94\\
mar       & 0.99 & 0.97 & \textbf{1.00} & 0.84 & 0.96 & 0.96 & \textbf{1.00} & \textbf{0.98} & 0.91 & 0.91 & 0.76 & 0.91 & \textbf{1.00} & 0.95 & 0.98 & \textbf{0.98} & 0.94 & 0.94\\
mni\_Beng & \textbf{1.00} & \textbf{1.00} & 0.99 & 0.99 & \textbf{1.00} & \textbf{1.00} & 0.99 & \textbf{1.00} & \textbf{1.00} & \textbf{1.00} & \textbf{1.00} & \textbf{1.00} & \textbf{1.00} & \textbf{1.00} & \textbf{1.00} & \textbf{1.00} & \textbf{1.00} & 0.99\\
mni\_Mtei & \textbf{1.00} & \textbf{1.00} & \textbf{1.00} & \textbf{1.00} & \textbf{1.00} & \textbf{1.00} & \textbf{1.00} & \textbf{1.00} & \textbf{1.00} & \textbf{1.00} & \textbf{1.00} & \textbf{1.00} & 0.91 & \textbf{1.00} & 0.69 & 0.71 & 0.99 & 0.98\\
npi       & 0.92 & 0.91 & 0.85 & 0.84 & 0.91 & 0.90 & 1.00 & 0.91 & 0.87 & 0.86 & 0.75 & 0.83 & 1.00 & 0.91 & \textbf{0.93} & \textbf{0.92} & 0.88 & 0.87\\
ory       & \textbf{1.00} & \textbf{1.00} & \textbf{1.00} & \textbf{1.00} & \textbf{1.00} & \textbf{1.00} & 0.92 & \textbf{1.00} & \textbf{1.00} & \textbf{1.00} & \textbf{1.00} & \textbf{1.00} & \textbf{1.00} & \textbf{1.00}  & \textbf{1.00} & \textbf{1.00} & \textbf{1.00} & \textbf{1.00}\\
pan       & \textbf{1.00} & \textbf{1.00} & 0.99 & 0.99 & \textbf{1.00} & \textbf{1.00} & \textbf{1.00} & \textbf{1.00} & \textbf{1.00} & \textbf{1.00} & 0.99 & \textbf{1.00} & 0.97 & \textbf{1.00}  & \textbf{1.00} &\textbf{1.00}& 0.96 & 0.97\\
san       & \textbf{1.00} & \textbf{1.00} & 0.95 & 0.95 & \textbf{1.00}& \textbf{1.00} & \textbf{1.00} & \textbf{1.00} & 0.96 & 0.97 & 0.60 & 0.96 & \textbf{1.00} & 0.97  & \textbf{1.00} & \textbf{1.00} & 0.98 & 0.98\\
sat       & \textbf{1.00}&\textbf{1.00} & \textbf{1.00} & \textbf{1.00} & \textbf{1.00}& \textbf{1.00} & \textbf{1.00} & \textbf{1.00} & \textbf{1.00} & \textbf{1.00}&\textbf{1.00}& 0.99 & 0.92 & \textbf{1.00} & 0.70 & 0.72 & 0.99 & 0.99\\
snd\_arab & \textbf{1.00} & \textbf{1.00}& 0.97 & 0.98 & 0.99 & \textbf{1.00} & \textbf{1.00}& \textbf{1.00} & 0.99 & \textbf{1.00} & 0.99 & 0.97 & 0.99 & 0.98  & \textbf{1.00} & \textbf{1.00} & \textbf{1.00}& 0.99\\
snd\_deva & 0.94 & 0.93 & 0.88 & 0.85 & 0.92 & 0.91 & 0.94 & 0.92 & 0.89 & 0.90 & 0.80 & 0.88 & 0.92 & 0.92  & 0.94 & 0.94 & 0.92 & 0.92\\
tam       & \textbf{1.00} & \textbf{1.00}& 0.99 & 0.99 & \textbf{1.00} & \textbf{1.00} & \textbf{1.00} & \textbf{1.00} & 0.99 & 0.99 & 0.90 &\textbf{1.00} & 0.99 & \textbf{1.00}  & \textbf{1.00} & \textbf{1.00} & 0.93 & 0.94\\
tel       & \textbf{1.00} & \textbf{1.00}& 0.99 & \textbf{1.00}& \textbf{1.00}& \textbf{1.00} &\textbf{1.00} & \textbf{1.00}& \textbf{1.00} & \textbf{1.00} & \textbf{1.00} & \textbf{1.00} &\textbf{1.00} & \textbf{1.00}  &\textbf{1.00} & \textbf{1.00} & 0.97 & 0.97\\
urd       & \textbf{1.00} & 0.98 & 0.93 & 0.93 & 0.97 & 0.97 & 1.00 & 0.99 & 0.96 & 0.96 & 0.97 & 0.97 & 0.98 & 0.98  & \textbf{1.00} & \textbf{1.00} & 0.81 & 0.79\\
\hline
\end{tabular}%
}
\caption{F1-Scores on ILID Dev and Test Sets for each classifier on each language. Highest scores in each language in respective datasets across models are marked in bold}
\label{tab:combined-f1-table}
\end{table*}

\bibliography{custom}
\bibliographystyle{acl_natbib}

\end{document}